\documentclass[lettersize,journal]{IEEEtran}
\usepackage{amsmath,amsfonts}
\usepackage{algorithmic}
\usepackage{algorithm}
\usepackage{array}
\usepackage[caption=false,font=footnotesize,labelfont=sf,textfont=sf]{subfig}
\usepackage{textcomp}
\usepackage{stfloats}
\usepackage{url}
\usepackage{verbatim}
\usepackage{graphicx}
\usepackage{cite}
\usepackage{amsmath,amssymb,booktabs}
\usepackage{color}
\usepackage{amsthm}
\usepackage{wasysym}

\newcommand{\bs}[1]{\boldsymbol {#1}}

\newcommand\doublecheck{\checkmark\kern-0.6em\checkmark}

\usepackage{bm}
\usepackage{braket}  
\usepackage{url}
\usepackage{multirow}
\usepackage{subcaption}
\begin{document}

\title{PolarBM: Complex-valued Boltzmann Machine for Modeling \\ Audio Signals in Polar and Log-polar Coordinates}

\author{Toru Nakashika, 
\IEEEmembership{Member, IEEE}, and Kohei Yatabe, \IEEEmembership{Member, IEEE}
\thanks{Manuscript received XXXXX XX, 202X; revised XXXXX XX, 202X... Date of current version XXXXX XX, 202X. The associate editor coordinating the review of this manuscript was XXXXX}
\thanks{T. Nakashika is with Graduate School of Informatics and Engineering, the University of Electro-Communications, Tokyo 182-8585, 
Japan (e-mail: nakashika@uec.ac.jp).}
\thanks{K. Yatabe is with the Department of Electrical Engineering and Computer Science, Tokyo University of Agriculture and Technology, Tokyo 184-8588, Japan (e-mail: yatabe@go.tuat.ac.jp).}
\thanks{Digital Object Identifier XXXXX}
\thanks{This work has been submitted to the IEEE for possible publication. Copyright may be transferred without notice, after which this version may no longer be accessible.} 
}

\markboth{IEEE Transactions on Audio, Speech and Language Processing, 
Vol. X, No. X, XXX 2026}
{T. Nakashika, K. Yatabe: 
PolarRBM: Complex-valued Boltzmann Machine for Modeling Audio Signals}


\maketitle

\begin{abstract}
Although vast amounts of data, such as audio signal spectra, are naturally represented using complex numbers, conventional machine learning methods often simplify complex-domain problems by employing frameworks designed for real-valued variables. While this simplification offers computational benefits, it discards structural information regarding the inherent relationship between amplitude and phase. In this paper, we propose a novel Boltzmann machine (BM), named \textit{PolarBM}, capable of naturally handling complex-valued variables in the polar coordinate (i.e., an amplitude-phase representation). PolarBM defines a probability density function for complex variables in which the phase explicitly depends on the amplitude, thereby capturing the physically important relationships of complex-valued signals. Furthermore, to process audio signals in accordance with human auditory perception, we propose \textit{LogPolarBM}, which models amplitude on a logarithmic scale. This extension yields a flexible conditional probability density function, a \textit{power-weighted noncentral complex Gaussian (PW-NCCG) distribution}, whose marginal amplitude distribution encompasses the Rice, Nakagami, and noncentral chi distributions as special cases. For practical applications, we also introduce the restricted variants of these proposed models: \textit{PolarRBM} and \textit{LogPolarRBM}. Experimental results demonstrate that by explicitly modeling the dependency between amplitude and phase, the proposed RBMs achieve superior modeling accuracy compared to conventional models, including deep neural networks. Although our experiments focus on audio signals, the utility of the proposed BMs is not limited to audio applications; their potential extends widely across various fields of science and engineering that involve complex-valued data, such as wireless communications and quantum mechanics.
\end{abstract}


\section{Introduction}
\IEEEPARstart{T}{he} Boltzmann machine (BM) \cite{ackley1985learning} has long served as a foundational model in the fields of stochastic neural networks and energy-based models. A BM defines the probability distribution of a system using an energy function to learn the underlying distribution of the data. This capability has enabled its application across various tasks, including feature extraction, representation learning, and data generation \cite{hinton2006reducing,salakhutdinov2009deep,mohamed2011acoustic,goodfellow2016deep}. However, despite its theoretical sophistication, the original BM is fundamentally constrained by its exclusive use of binary variables. While this binary representation is sufficient for certain applications \cite{salakhutdinov2007restricted,hinton2006fast}, it presents a critical limitation when modeling real-world physical data. Such data are inherently continuous and often exhibit complex dependencies that cannot be adequately captured by binary variables. To overcome this limitation, various extensions of the BM have been proposed \cite{courville2011spike}. For instance, the Gaussian BM \cite{Lee:2008uz} can model real-valued data, while the gamma BM \cite{nakashika2021gamma} is designed for positive-valued data.

In physics and engineering, vast amounts of data, such as the spectra of audio and wireless signals, are most naturally represented using the complex number system. However, few existing models adequately handle the algebraic and geometric properties of complex variables. Conventional machine learning methods often simplify the problem by completely discarding phase information \cite{lu2013speech} or by treating the real and imaginary parts as independent real-valued variables \cite{bassey2021survey}. Although several BM extensions have been proposed to handle complex variables, these approaches often rely on restrictive simplifications. For instance, in the directional-unit BM (DUBM) \cite{zemel1993directional}, only the phase is treated as a random variable, while the amplitude is strictly fixed to 1. The complex amplitude-phase Boltzmann machine (CAPBM) \cite{li2020complex} allows the amplitude to vary, but restricts it to binary values (0 or 1). The complex-valued restricted Boltzmann machine (complex-valued RBM) \cite{nakashika2018complex} models correlations between the real and imaginary components, rather than the more physically meaningful amplitude-phase representation. Furthermore, the Gaussian von Mises RBM (GVM-RBM) \cite{nakashika2025gamma} employs a gamma distribution for the amplitude and a von Mises distribution for the phase, but inherently assumes mutual independence between the two. While these simplifications offer computational convenience, they artificially decouple the intrinsic mathematical relationship between amplitude and phase, leading to the loss of critical structural information inherent to the signals.


To bridge this gap, in this paper, we propose the \textit{polar Boltzmann machine (PolarBM)}, a natural extension of the standard BM capable of directly modeling complex variables. Unlike conventional methods that coerce complex data into real-valued frameworks, our approach formulates the model fundamentally within the complex domain. The PolarBM generalizes standard BMs while successfully capturing the physically significant dependencies between amplitude and phase. By explicitly modeling this relationship, the proposed complex distribution offers deeper insights into the statistical behavior of complex signals. Furthermore, we introduce the \textit{polar restricted Boltzmann machine (PolarRBM)}, which incorporates hidden units into this framework. Featuring a bipartite topology with connections exclusively between complex-valued visible units and binary hidden units, the PolarRBM enables the extraction of latent representations from complex data, thereby facilitating downstream tasks such as data generation and classification.

In addition to the PolarBM, we propose the \textit{LogPolarBM}, which models amplitude on a logarithmic scale. In the field of audio processing, amplitude spectra are frequently represented logarithmically to align with the perception of the human auditory system. While such logarithmic representations of amplitude have proven important for BMs \cite{nakashika2021gamma}, they have conventionally been modeled independently from phase. To achieve a more natural and accurate representation of complex spectral structures, we extend the PolarBM and introduce the LogPolarBM that incorporates a log-amplitude energy term analogous to that of the gamma BM \cite{nakashika2021gamma}. This extension yields a highly flexible conditional probability density function for each unit, namely a \textit{power-weighted noncentral complex Gaussian (PW-NCCG) distribution} \cite{nakashika2026power}. The marginal amplitude distribution of each unit generalizes several well-known distributions, encompassing not only the gamma distribution but also the Rice \cite{rice1945mathematical}, Nakagami \cite{nakagami1960m}, and noncentral chi distributions \cite{johnson1995continuous}. Finally, to facilitate practical applications, we also introduce its restricted counterpart, the \textit{LogPolar restricted Boltzmann machine (LogPolarRBM)}, along with an approximation of the expected value of PW-NCCG to enable computationally efficient inference.

The remainder of this paper is organized as follows. Section~\ref{sec:Preliminaries} briefly reviews existing BMs relevant to this study. Sections~\ref{subsec:PolarBM} and \ref{subsec:LogPolarBM} propose the PolarBM and LogPolarBM, respectively, while Section~\ref{subsec:RelatedDistribution} discusses the relationships between the proposed distributions and conventional models. Section~\ref{sec:proposedRBMs} introduces the RBM variants of the proposed models, their training objectives, and optimization procedures. Section~\ref{sec:exp} evaluates the proposed RBMs through experiments, and Section~\ref{sec:conclusion} concludes the paper.

\section{Preliminaries}
\label{sec:Preliminaries}

\subsection{Boltzmann Machine (BM)}

BM~\cite{ackley1985learning} is a probabilistic model that can effectively capture the fundamental essence of systems in diverse fields of science and engineering.
For a $D$-dimensional binary input feature $\bs{x} \in \mathbb{B}^D$, BM provides the following probability density:
\begin{align}
p(\bs{x}) &= \frac{1}{Z}\mathrm{e}^{-E_\mathrm{BM}(\bs{x})}, \\
    E_\mathrm{BM}(\bs{x}) &= - \frac{1}{2} \bs{x}^{\sf T} \mathbf{W} \bs{x} - \bs{b}^{\sf T} \bs{x},
    \label{eq:defBMenergy}
\end{align}
where $(\cdot)^{\sf T}$ represents the transpose, $Z = \int p(\bs{x}) \,\mathrm{d} \bs{x} $ is the normalization term, $\mathbf{W} \in \mathbb{R}^{D \times D}$ ($\mathbf{W}^{\sf T} = \mathbf{W}, \ {\rm diag}(\mathbf{W}) = \bs{0}$) is the weight matrix encoding the strength of connectivity between each unit, and $\bs{b} \in \mathbb{R}^D$ is the bias parameter for each unit.
From this definition, the (conditional) probability density for the $d$th unit $x_d$ can be computed using the other units $\bs{x}_{\not d} = [x_1, \cdots, x_{d-1}, x_{d+1}, \cdots, x_D]^{\sf T}$ as the following Bernoulli distribution:
\begin{align}
p(x_d \, | \, \bs{x}_{\not d}) &= \mathcal{B}\bigl(x_d; \bs{\varsigma}(\mathbf{W}_{\!\not d,d}^{\sf T} \, \bs{x}_{\not d} + b_d)\bigr), 
\end{align}
where $\mathcal{B}(\,\cdot\,; \pi)$ represents the Bernoulli distribution with success rate $\pi$, $\bs{\varsigma}(\cdot)$ denotes the sigmoid function, and $\mathbf{W}_{\!\not d,d}\in \mathbb{R}^{D-1}$ is the $(D-1)$-dimensional vector obtained by removing the $d$th row and selecting the $d$th column from $\mathbf{W}$.

\subsection{Gaussian BM (GaussBM)}

The above BM representing binary variables has been extended to represent real values.
Since merely extending the domain from $\mathbb{B}^D$ to $\mathbb{R}^D$ causes the probability density to diverge, an inverse variance $\bs{\beta}$ (such that ${\rm diag}(\mathbf{W}) = -2\bs{\beta}$, $\beta_d \in \mathbb{R}_{>0} $) is introduced, and a decay term ${\rm e}^{-\beta_dx_d^2}$ is added to each unit.
This constitutes the Gaussian BM (GaussBM) \cite{Lee:2008uz}.
While the energy function remains the same as Eq.~\eqref{eq:defBMenergy}, the domain of $\bs{x}$ and the definition of $\mathbf{W}$ differ.
The conditional probability density of a given unit $x_d$ follows a Gaussian distribution as follows:
\begin{align}
p(x_d \, | \, \bs{x}_{\not d}) &= \mathcal{N}\biggl(x_d; \frac{\mathbf{W}_{\!\not d,d}^{\sf T} \, \bs{x}_{\not d} + b_d}{2\beta_d}, \frac{1}{2\beta_d}\biggr), 
\end{align}
where $\mathcal{N}(\,\cdot\,; \mu, \sigma^2)$ represents the Gaussian distribution with mean $\mu$ and variance $\sigma^2$.

\subsection{Directional-unit BM (DUBM)}

The directional-unit Boltzmann machine (DUBM)~\cite{zemel1993directional} is a variant of BM for handling phase of a complex variable.
In DUBM, $D$-dimensional unimodular complex-valued input feature $\bs{z} \in \mathbb{C}^D$ ($\forall d$, $|z_d| = 1$, i.e., $z_d = \mathrm{e}^{\rm i \theta_d}$, $\theta_d \in [-\pi, \pi)$) is represented by the following probability density:
\begin{align}
p(\bs{z}) &= \frac{1}{Z}\mathrm{e}^{-E_\mathrm{DUBM}(\bs{z})}, \\
    E_\mathrm{DUBM}(\bs{z}) &= - \frac{1}{2} \bs{z}^{\sf H} \mathbf{W} \bs{z} - \Re \left[ \bs{b}^{\sf H} \bs{z} \right],
    \label{eq:defDUBMenergy}
\end{align}
where $\mathrm{i}$ is the imaginary unit, $\Re[\cdot]$ represents the real part, $(\cdot)^{\sf H}$ is the Hermitian transpose, $Z = \int p(\bs{z}) \, \mathrm{d} \bs{z} $ is the normalization term, and $\mathbf{W} \in \mathbb{C}^{D \times D}$ ($\mathbf{W}^{\sf H} = \mathbf{W}, \ {\rm diag}(\mathbf{W}) = \bs{0}$) and $\bs{b} \in \mathbb{C}^D$ are the model parameters.
The conditional probability density of a unimodular complex-valued unit $z_d$ with phase $\theta_d$ can be derived as follows:
\begin{align}
p(z_d \, | \, \bs{z}_{\not d}) &= p(\theta_d \, | \, \bs{z}_{\not d}) \\
&= \mathcal{VM}\bigl(\theta_d; \angle(\mathbf{W}_{\!\not d,d}^{\sf H} \, \bs{z}_{\not d} + b_d), |\mathbf{W}_{\!\not d,d}^{\sf H} \, \bs{z}_{\not d} + b_d|\bigr), \nonumber
\end{align}
where $\mathcal{VM}(\,\cdot\,; \mu, \kappa)$ is the von Mises distribution with mean phase $\mu\in[-\pi,\pi)$ and concentration parameter $\kappa\in\mathbb{R}_{>0}$, $|\cdot|$ is the absolute value, and $\angle(\cdot)$ is the principal argument.

\subsection{Gamma BM (GammaBM)}
The gamma Boltzmann machine (GammaBM) \cite{nakashika2021gamma} is an extension of BM designed to model positive-valued variables.
In GammaBM, the probability distribution of a $D$-dimensional positive vector $\bs{x} \in \mathbb{R}_{>0}^D$ is defined through the following energy function:
\begin{align}
    E_{\Gamma}(\bs{x}) = &- \frac{1}{2} \bs{x}^\mathsf{T} \mathbf{W} \bs{x}  - \bs{b}^\mathsf{T} \bs{x}\nonumber\\
    &- \frac{1}{2} \log (\bs{x})^\mathsf{T} \mathbf{V} \log (\bs{x})  - \bs{c}^\mathsf{T} \log (\bs{x}),
    \label{eq:defGammaBoltzmannMachine}
\end{align}
where $\log(\cdot)$ is the element-wise logrithmic function, $-\mathbf{W}\in\mathbb{R}^{D\times D}_{>0}$ ($\mathbf{W}^{\sf T\!} = \mathbf{W}$, ${\rm diag}(\mathbf{W})=\bs{0}$), $\mathbf{V}\in\mathbb{R}^{D\times D}_{>0}$ ($\mathbf{V}^{\sf T\!} = \mathbf{V}$, ${\rm diag}(\mathbf{V})=\bs{0}$), $-\bs{b}\in\mathbb{R}^D_{\ge 0}$, and $\bs{c}+1\in\mathbb{R}^D_{\ge 0}$.
Under this formulation, the conditional probability density of each unit $x_d$ follows a gamma distribution:
\begin{align}
p(x_d \, | \, \bs{x}_{\not d}) &= \mathcal{G}\bigl(x_d; \mathbf{V}_{\!\not d,d}^{\sf T} \log(\bs{x}_{\not d}) + c_d+1, \mathbf{W}_{\!\not d,d}^{\sf T} \, \bs{x}_{\not d} + b_d\bigr),
\end{align}
where $\mathcal{G}(\,\cdot\,;\alpha,\beta)$ is the gamma distribution with the shape parameter $\alpha$ and rate $\beta$.

As in Eq.~\eqref{eq:defGammaBoltzmannMachine}, GammaBM handles the variable in both the linear scale, via $\bs{x}^\mathsf{T} \mathbf{W} \bs{x}$, and the logarithmic scale, via $\log (\bs{x})^\mathsf{T} \mathbf{V} \log (\bs{x})$.
In other words, within a single energy function, it provides a unified modeling framework that captures not only absolute changes in values via linear interactions but also relative, ratio-based scaling via logarithmic interactions.
This property of GammaBM is advantageous for audio applications \cite{nakashika2021gamma} because human perception handles magnitude of sound in approximate logarithmic scale.

\begin{figure*}[t]
\centerline{\includegraphics[width=1.4\columnwidth]{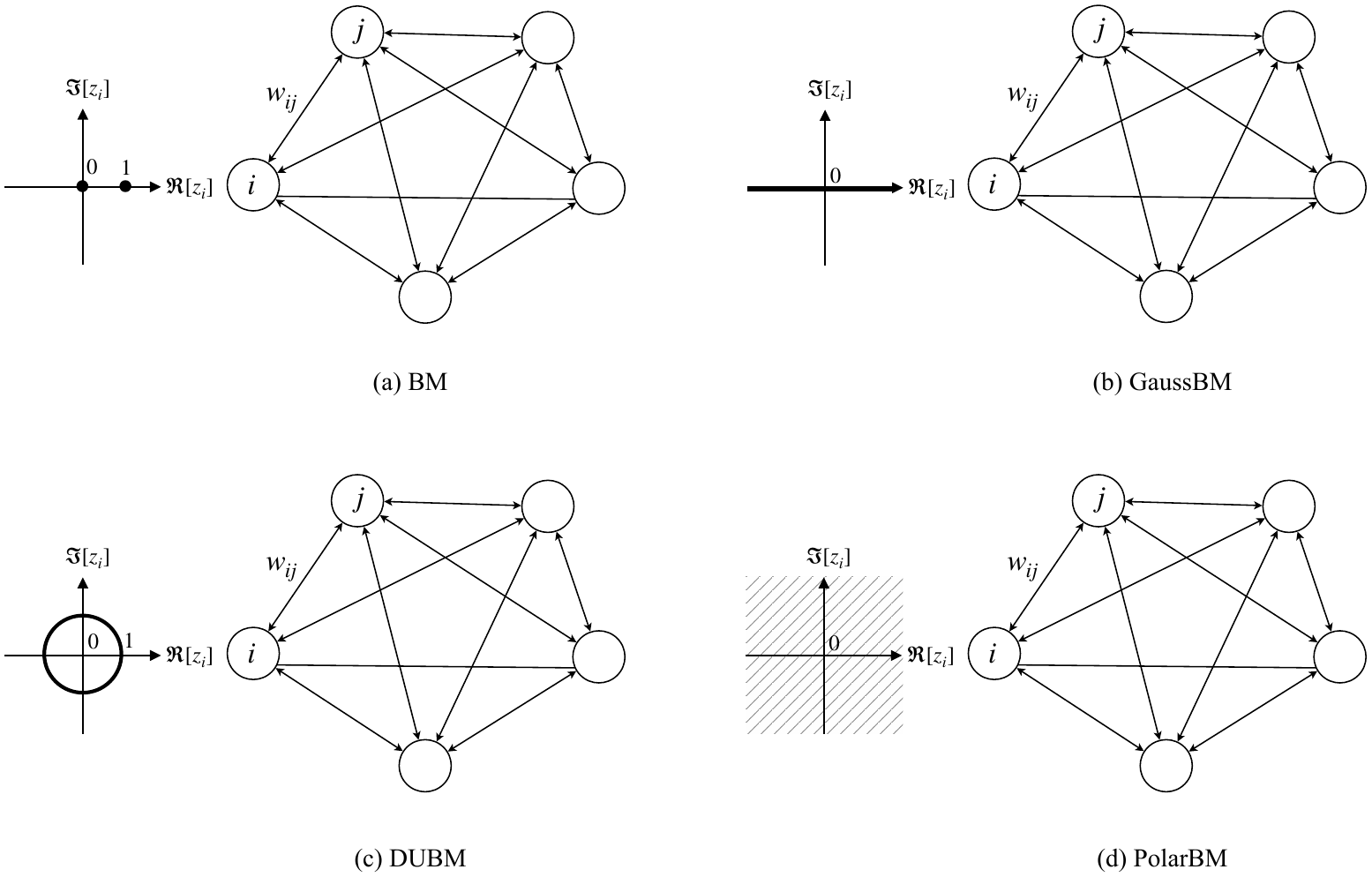}}
\caption{Graphical representation of (a) standard BM, (b) GaussBM, (c) DUBM, and (d) proposed PolarBM.}
\label{fig:PolarBM}
\end{figure*}

\section{PolarBM and Its Properties}

Complex numbers are essential for modeling and analyzing real-world signals and systems.
However, the standard BMs can only handle specific types of variables, e.g., GaussBM for real numbers and DUBM for unimodular complex numbers.
In this study, we propose a novel BM that can naturally handle complex numbers in the polar coordinate.

\subsection{PolarBM}
\label{subsec:PolarBM}

Upon examining the definitions of BM and DUBM in Eqs.~\eqref{eq:defBMenergy} and \eqref{eq:defDUBMenergy}, it appears that arbitrary complex variables $\bs{z} \in \mathbb{C}^D$ can be handled by adopting the following definition:
\begin{align}
z_d &= r_d \, \mathrm{e}^{{\rm i} \theta_d}, \quad r_d \in \mathbb{R}_{>0}, \quad \theta_d \in [-\pi, \pi), \\
p(\bs{z}) &= \frac{1}{Z}\mathrm{e}^{-E(\bs{z})}, \\
    E(\bs{z}) &= - \frac{1}{2} \bs{z}^{\sf H} \mathbf{W} \bs{z} - \Re \left[ \bs{b}^{\sf H} \bs{z} \right],
\end{align}
When $\mathbf{W}^{\sf H} = \mathbf{W}$ and ${\rm diag}(\mathbf{W}) = \bs{0}$ ($\mathbf{W} \in \mathbb{C}^{D \times D}$), the conditional probability density of a unit $z_d$ with given $\bs{z}_{\not d}$ is
\begin{align}
p(z_d \, | \, \bs{z}_{\not d}) &= \frac{1}{Z}\mathrm{e}^{\frac{1}{2} \bs{z}^{\sf H} \mathbf{W} \bs{z} + \Re \left[ \bs{b}^{\sf H} \bs{z} \right]} \nonumber \\
&= \frac{1}{Z}\mathrm{e}^{\frac{1}{2} \sum_{i=1}^D \sum_{j=1}^D \overline{z}_i w_{ij} z_{j} + \Re \left[ \sum_{i=1}^D \overline{b}_iz_i \right]} \nonumber \\
&= \frac{1}{Z'}\mathrm{e}^{\frac{1}{2} \sum_{i=1}^D (\overline{z}_i w_{id} z_d + \overline{z}_d \overline{w}_{id} z_i) +\Re \left[ \overline{b}_dz_d \right]} \nonumber \\
&= \frac{1}{Z'}\mathrm{e}^{\frac{1}{2} (\overline{\mathbf{W}_{\!\not d,d}^{\sf H} \bs{z}_{\not d}} \, z_d + \mathbf{W}_{\!\not d,d}^{\sf H} \bs{z}_{\not d} \, \overline{z}_d) +\Re \left[ \overline{b}_dz_d \right]} \nonumber \\
&= \frac{1}{Z'}\mathrm{e}^{ \Re \left[ \overline{\mathbf{W}_{\!\not d,d}^{\sf H} \bs{z}_{\not d}} \, z_d \right] +\Re \left[ \overline{b}_dz_d \right]} \nonumber \\
&= \frac{1}{Z'}\mathrm{e}^{ \Re \left[ (\overline{\mathbf{W}_{\!\not d,d}^{\sf H} \bs{z}_{\not d} + b_d}) z_d \right]},
\end{align}
where $Z'$ is the normalization constant independent of $z_d$.
Let $u_d = \mathbf{W}_{\!\not d,d}^{\sf H} \, \bs{z}_{\not d} + b_d = a_d \, {\rm e}^{{\rm i} \phi_d}$, $a_d \in \mathbb{R}_{>0}$, $\phi_d \in [-\pi, \pi)$, then
\begin{align}
p(z_d \, | \, \bs{z}_{\not d}) &= \frac{1}{Z'}\mathrm{e}^{ \Re \left[ \overline{u}_d z_d \right]} \\
&= \frac{1}{Z'}\mathrm{e}^{ a_d |z_d| \cos(\angle z_d-\phi_d)}.
\end{align}
The distribution of magnitude $r_d = 
|z_d|$ can be obtained by marginalizing phase $\theta_d=\angle z_d$ in the polar coordinate system, with the variable transformation $p(z)\, \mathrm{d}z=p(r,\theta)\,r\,\mathrm{d}r\,\mathrm{d}\theta$, as
\begin{align}
p(r_d \, | \, \bs{z}_{\not d}) &= \int_{-\pi}^\pi p(z_d \, | \, \bs{z}_{\not d}) \, r_d \, \mathrm{d}\theta_d \\
&= \frac{r_d}{Z'} \int_{-\pi}^\pi \mathrm{e}^{ a_d r_d \cos(\theta_d-\phi_d)} \,\mathrm{d}\theta_d \\
&= \frac{2\pi}{Z'} r_d \, I_0(a_d \, r_d),
\end{align}
where $I_0(x)$ is the modified Bessel function of the first kind of order zero.
However, this probability density diverges when $r_d \rightarrow\infty$ as $r_d \, I_0(a_d \, r_d) \rightarrow\infty$.

To introduce decay similar to GaussBM, we propose to restrict ${\rm diag}(\mathbf{W}) = -2\bs{\beta}$ with $\beta_d >0$ $(\forall d)$.
This modification adds $-\beta_d \, \overline{z}_d \, z_d = -\beta_d \, |z_d|^2$ to the exponent of the energy function, and the conditional probability density of $z_d$ becomes
\begin{align}
p(z_d \, | \, \bs{z}_{\not d}) &= \frac{1}{Z'}\mathrm{e}^{- \beta_d |z_d|^2 + a_d |z_d| \cos(\angle z_d-\phi_d)}. \label{eq:pz}
\end{align}
Its marginalization over the phase $\theta_d$ gives the following distribution for the magnitude $r_d$,
\begin{align}
p(r_d \, | \, \bs{z}_{\not d}) &= \frac{r_d}{Z'} \int_{-\pi}^\pi \mathrm{e}^{-\beta_d r_d^2 + a_d r_d \cos(\theta_d-\phi_d)} \, \mathrm{d}\theta_d \\
&\propto r_d \, \mathrm{e}^{-\beta_d r_d^2} I_0(a_d \, r_d) \\
&\propto {\rm Rice} \Bigl(r_d; \frac{a_d}{2\beta},\frac{1}{\beta}\Bigr),
\end{align}
where ${\rm Rice} (\,\cdot\,; \nu,\sigma^2)$ is the Rician distribution%
\footnote{Note that the Rician distribution is usually defined as ${\rm Rice} (r; \nu,\sigma^2)=\frac{r}{\sigma^2}{\rm e}^{-{(r^2+\nu^2)}/{2\sigma^2}} I_0(\frac{r\nu}{\sigma^2})$.
In this paper, to focus on the complex number system, we define it as ${\rm Rice} (r; \nu,\sigma^2)=\frac{2r}{\sigma^2}{\rm e}^{-{(r^2+\nu^2)}/{\sigma^2}} I_0(\frac{2r\nu}{\sigma^2})$ so that the variance of a complex-valued variable $z$ becomes $\sigma^2$ (i.e., the variance $\sigma^2$ is $1/2$ of that in the usual definition).}
with the noncentrality parameter $\nu$ and variance $\sigma^2$.
Moreover, when the magnitude is known, the conditional probability distribution of $z_d$ becomes the von Mises distribution as follows:
\begin{align}
p(\theta_d \, | \, r_d, \, \bs{z}_{\not d}) &\propto \mathrm{e}^{a_d r_d \cos(\theta_d-\phi_d)} \\
&\propto \mathcal{VM}\bigl(\theta_d; \phi_d, a_d \, r_d\bigr), \label{eq:pangle_r}
\end{align}

As in Eqs.~\eqref{eq:pz} and \eqref{eq:pangle_r}, the phase and magnitude are not independent; instead, the concentration parameter of phase is proportional to the magnitude $r_d$.
This is consistent with the intuition regarding the complex spectrum of a speech signal.
The phase spreads uniformly (becoming uncertain) when the magnitude is small, such as in silent intervals, whereas the phase becomes deterministic when a strong magnitude is observed, such as in voiced intervals. That is, a probabilistic model for complex-valued variables suitable for audio signals can be naturally obtained by this formulation.
We propose BM based on Eq.~\eqref{eq:pz} and name it \textit{PolarBM} to capture the relationships among complex-valued units, where phase and amplitude are interdependent.

In summary, the proposed PolarBM is defined as
\begin{align}
z_d &= r_d \, \mathrm{e}^{{\rm i} \theta_d}, \quad r_d \in \mathbb{R}_{>0}, \quad \theta_d \in [-\pi, \pi), \\
p(\bs{z}) &= \frac{1}{Z}\mathrm{e}^{-E_{\rm Polar}(\bs{z})}, \\
    E_{\rm Polar}(\bs{z}) &= - \frac{1}{2} \bs{z}^{\sf H} \mathbf{W} \bs{z} - \Re \left[ \bs{b}^{\sf H} \bs{z} \right], \label{eq:eulerbm1}
\end{align}
where $Z = \int p(\bs{z}) \, \mathrm{d} \bs{z} $ is the normalization term, and the model parameters are $\mathbf{W} \in \mathbb{C}^{D \times D}$ ($\mathbf{W}^{\sf H} = \mathbf{W}$, ${\rm diag}(\mathbf{W}) = -2\bs{\beta}$, $\beta_d > 0$) and $\bs{b} \in \mathbb{C}^D$.
Fig.~\ref{fig:PolarBM} illustrates the differences between the proposed PolarBM and other existing BMs, including binary BM, GaussBM, and DUBM.
While all the models describe interactions between two units, they differ in the respective domains.
Note that, when $r_d=1$ and $\beta_d \rightarrow 0$ ($\forall d$), then PolarBM becomes $p(\theta_d \, | \, r_d=1, \, \bs{z}_{\not d}) = \mathcal{VM}(\theta_d; \alpha_d, a_d)$, which is the same as DUBM.
Moreover, when all of $\bs{z}$, $\mathbf{W}$ and $\bs{b}$ are real-valued, then PolarBM coincides with GaussBM.
Therefore, the proposed PolarBM can be regarded as a natural extention of the existing BMs.

\begin{figure*}[t]
\centerline{\includegraphics[width=2.0\columnwidth]{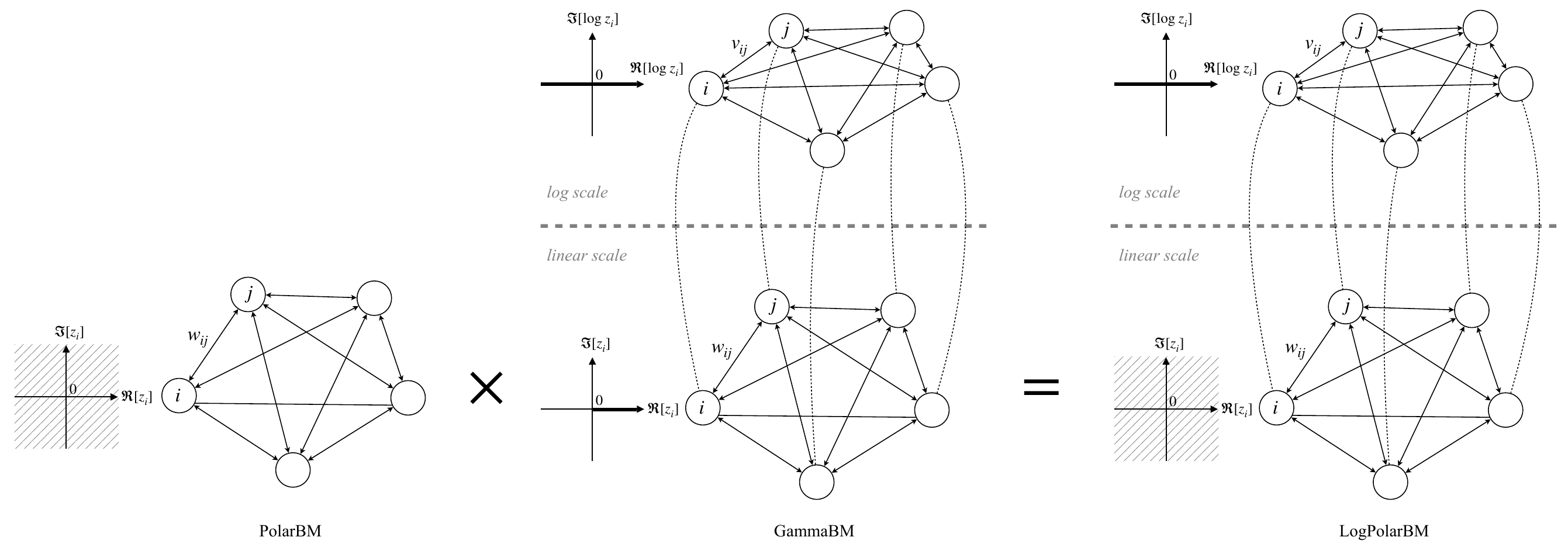}}
\caption{Conceptual diagram of the proposed LogPolarBM, which combines the complex-valued representation of PolarBM with the log-amplitude representation capability of GammaBM to capture diverse audio characteristics.}
\label{fig:logpolarbm}
\end{figure*}

\subsection{LogPolarBM}
\label{subsec:LogPolarBM}

The conditional probability density of the unit $z_d$ in Eq.~\eqref{eq:pz} can be interpreted as follows.
When Eq.~\eqref{eq:pz} is expressed using the Cartesian coordinate system $z_d = x_d + {\rm i} y_d$ ($x_d, \, y_d \in \mathbb{R}$), it can be rewritten as a circularly symmetric (isotropic) complex normal distribution as follows:
\begin{align}
p(z_d \, | \, \bs{z}_{\not d}) &\propto \mathrm{e}^{- \beta_d r_d^2 + a_d r_d (\cos \theta_d \cos \phi_d + \sin \theta_d \sin \phi_d)} \nonumber \\
&= \mathrm{e}^{- \beta_d (x_d^2 + y_d^2) + \Re[u_d]x_d + \Im[u_d]y_d} \\
&\propto \mathrm{e}^{- \beta_d \left[ \left(x_d - \frac{\Re[u_d]}{2\beta_d}\right)^2 + \left(y_d - \frac{\Im[u_d]}{2\beta_d}\right)^2\right]} \\
&\propto \mathrm{e}^{- \beta_d \left|z_d-\frac{u_d}{2\beta_d}\right|^2} \\
&\propto \mathcal{N}_{\mathrm{c}}\Bigl(z_d; \frac{u_d}{2\beta_d}, \frac{1}{\beta_d}\Bigr),
\end{align}
where the complex normal distribution with mean $\mu \in \mathbb{C}$ and variance $\sigma^2 \in \mathbb{R}_{>0}$ is given by
\begin{align}
\mathcal{N}_{\mathrm{c}}(z; \mu,\sigma^2) = \frac{1}{\pi \sigma^2} {\rm e}^{- \frac{|z-\mu|^2}{\sigma^2}}.
\end{align}
This indicates that PolarBM proposed in the previous subsection handles only linear relationships between the units.
However, such a simple model may not be suitable for representation of audio signals (such as speech) because the human auditory system perceives the magnitude of audio signals with a logarithmic-like scale.

In this subsection, we further propose LogPolarBM, a variant of PolarBM integrated with GammaBM for the logarithmic representation of magnitude spectra, as illustrated in Fig.~\ref{fig:logpolarbm}.
We define the LogPolarBM as
\begin{align}
z_d =& r_d \, \mathrm{e}^{{\rm i} \theta_d}, \quad r_d \in \mathbb{R}_{>0}, \quad \theta_d \in [-\pi, \pi), \\
p(\bs{z}) =& \frac{1}{Z}\mathrm{e}^{-E_{\rm LogPolar}(\bs{z})}, \\
    E_{\rm LogPolar}(\bs{z}) =& - \frac{1}{2} \bs{z}^{\sf H} \mathbf{W} \bs{z} - \Re \left[ \bs{b}^{\sf H} \bs{z} \right] \nonumber\\
    &- \frac{1}{2} \log|\bs{z}|{^{\sf T}} \mathbf{V} \log|\bs{z}| - \bs{c}^{\sf T} \log |\bs{z}| , \label{eq:eulerbm2}
\end{align}
where $\mathbf{V} \in \mathbb{R}^{D \times D}$ ($\mathbf{V}^{\sf T} = \mathbf{V}$, ${\rm diag}(\mathbf{V}) = \bs{0}$) and $\bs{c} \in \mathbb{R}^D$ are the parameters related to the logarithmic magnitude spectra, which must satisfy $ \mathbf{V}_{\!\not d,d}^{\sf T}\log|\bs{z}_{\not d}| + c_d > -2$ ($\forall d$) for $\mathbf{V}_{\!\not d,d}\in \mathbb{R}^{D-1}$.
This is a natual combination of PolarBM in Eq.~\eqref{eq:eulerbm1} and GammaBM in Eq.~\eqref{eq:defGammaBoltzmannMachine}.
Compared to PolarBM in Eq.~\eqref{eq:eulerbm1}, this LogPolarBM adds logarithmic amplitude terms.
By employing this logarithmic representation, LogPolarBM effectively captures perceptually relevant features, consistent with human auditory perception.
Note that PolarBM in Eq.~\eqref{eq:eulerbm1} is the special case of LogPolarBM in Eq.~\eqref{eq:eulerbm2} when $\mathbf{V}=\mathbf{O}$ and $\bs{c}=\bs{0}$ ($\forall d$, $\mathbf{V}_{\!\not d,d}^{\sf T}\log|\bs{z}_{\not d}| + c_d = 0$).

For LogPolarBM, the conditional probability density of a unit $z_d$ can be written as
\begin{align}
p(z_d|\bs{z}_{\not d}) &\propto |z_d|^{2\alpha_d-2} \, \mathrm{e}^{- \beta_d |z_d|^2 + a_d |z_d| \cos(\angle z_d-\phi_d)} \\
&\propto |z_d|^{2\alpha_d-2} \, \mathrm{e}^{- \beta_d \left|z_d-\frac{u_d}{2\beta_d}\right|^2} \\
&\propto \mathcal{N}_{\mathrm{pw}}\Bigl(z_d; \frac{u_d}{2\beta_d}, \frac{1}{\beta_d}, \alpha_d\Bigr) \label{eq:pzz_},
\end{align}
where $\alpha_d = (\mathbf{V}_{\!\not d,d}\log|\bs{z}_{\not d}| + c_d)/2 + 1$ ($>0$), and $\mathcal{N}_{\mathrm{pw}}(z)$ is the following distribution termed the \textit{power-weighted noncentral complex Gaussian (PW-NCCG) distribution}:
\begin{align}
\mathcal{N}_{\mathrm{pw}}\bigl(z; \mu, \sigma^2, \alpha\bigr) = \frac{|z|^{2\alpha-2} \, \mathrm{e}^{- \frac{|z-\mu|^2}{\sigma^2}}}{\pi \, \sigma^{2\alpha} \,\Gamma(\alpha)\,L_{\alpha-1}(-\frac{|\mu|^2}{\sigma^2})}, \label{eq:defPWN}
\end{align}
where $\Gamma(\cdot)$ is the gamma function, $L_n(\cdot)$ is the $n$th-order Laguerre function, and $\mu \in \mathbb{C}$, $\sigma^2 \in \mathbb{R}_{>0}$ and $\alpha \in \mathbb{R}_{>0}$ are the mean (center of gravity), variance, and degree parameters, respectively.
The degree parameter $\alpha$ governs the shape of the distribution (note that PW-NCCG reduces to the complex normal distribution when $\alpha = 1$, i.e., $\mathcal{N}_{\mathrm{pw}}(z; \mu, \sigma^2,1) = \mathcal{N}_{\mathrm{c}}(z; \mu, \sigma^2)$).
In particular, a value of $0<\alpha<1$ concentrates the probability density around the origin.
This is an advantageous property for representing spectra of sparse signals such as vowels, as it can capture sharp spectral peaks with a heavier-tailed distribution.
On the other hand, a value of $\alpha>1$ suppresses the density around the origin.
In this case, it well represents spectra of more complex signals such as fricative sounds that tend to exhibit more diffuse energy distributions, as it approaches a Gaussian-like distribution with a lighter tail.
Therefore, LogPolarBM (based on PW-NCCG) provides greater flexibility in modeling a wider range of audio signals than the standard complex normal distribution-based PolarBM, thanks to the additional parameter $\alpha$.
Some more details on PW-NCCG can be found in \cite{nakashika2026power}.

Similar to that of PolarBM, given the magnitude $r_d$, the conditional probability density of a unit $z_d$ for LogPolarBM is the following von Mises distribution:
\begin{align}
p(\theta_d|r_d,\bs{z}_{\not d}) &\propto \mathrm{e}^{a_d r_d \cos(\theta_d-\phi_d)} \\
&\propto \mathcal{VM}\bigl(\theta_d; \phi_d, a_d \, r_d\bigr).
\end{align}
On the other hand, the distribution of magnitude, $p(r_d|\bs{z}_{\not d})$, is calculated as follows:
\begin{align}
p(r_d|\bs{z}_{\not d}) &= \int_{-\pi}^{\pi} \mathcal{N}_{\mathrm{pw}}\Bigl(z_d; \frac{u_d}{2\beta_d}, \frac{1}{\beta_d}, \alpha_d\Bigr) \, \mathrm{d}\theta \\
&= p\Bigl(r_d; \frac{u_d}{2\beta_d}, \frac{1}{\beta_d}, \alpha_d\Bigr),
\label{eq:NakashikaDistribution}
\end{align}
where the last term is given by
\begin{align}
p(r;\nu,\sigma^2,\alpha) = \frac{2\,r^{2\alpha-1} \,
\mathrm{e}^{- \frac{r^2+\nu^2}{\sigma^2}}
\, I_0\!\left(\frac{2 \nu r}{\sigma^2}\right)}
{\sigma^{2\alpha} \, \Gamma(\alpha) \, L_{\alpha-1}\!\left(-\frac{\nu^2}{\sigma^2}\right)}.
\label{eq:pr}
\end{align}
This is an extension of the Rician distribution that can be derived from PolarBM (when $\alpha=1$, it coincides with the Rician distribution; see Eq.~(9) in~\cite{nakashika2026power}).
This distribution can also be interpreted as a generalization of several other magnitude or power distributions, highlighting the versatility and effectiveness of LogPolarBM based on this distribution.
We will summarize these relationships in the next subsection.

\subsection{Relationship of Eq.~\eqref{eq:pr} to Existing Distributions}
\label{subsec:RelatedDistribution}

The magnitude distribution in Eq.~\eqref{eq:pr} is related to the other standard distributions as follows.

When $\nu=0$, $\sigma^2=\frac{\Omega}{\alpha}$ and $\Omega>0$, Eq.~\eqref{eq:pr} becomes the Nakagami distribution as follows:
\begin{align}
p\Bigl(r;0,\frac{\Omega}{\alpha},\alpha\Bigr) &= \frac{2 \, \alpha^\alpha  \, r^{2\alpha-1} \, \mathrm{e}^{- \frac{\alpha}{\Omega}r^2} }{\Omega^{\alpha} \, \Gamma(\alpha)} \\
&= {\rm Nakagami}\bigl(r; \alpha,\Omega\bigr).
\end{align}
Therefore, Eq.~\eqref{eq:pr} can be viewed as a generalization of the Nakagami distribution.
Although the distribution in Eq.~\eqref{eq:pr} is naturally derived from the energy function in Eq.~\eqref{eq:eulerbm2}, these relationships to the Rice and Nakagami distributions suggest its appropriateness, given their widespread application in Telecommunications Engineering.

When $\sigma^2=2$, $\alpha=\frac{k}{2}$ and $k>0$, Eq.~\eqref{eq:pr} can be rewritten as the noncentral chi distribution:
\begin{align}
p\Bigl(r;\nu,2,\frac{k}{2}\Bigr) &= \frac{2^{1-\frac{k}{2}} \, r^{k-1} \, \mathrm{e}^{- \frac{r^2+\nu^2}{2}} \, I_0(\nu \, r) }{\Gamma(\frac{k}{2}) \, L_{\frac{k}{2}-1}(-\frac{\nu^2}{2})} \label{eq:chi_like}\\
&\propto r^{k-1} \, \mathrm{e}^{- \frac{r^2}{2}} \, I_0(\nu \, r),
\end{align}
where the noncentral chi distribution is given by
\begin{align}
\chi' (r;k,\nu) &= \nu^{1-\frac{k}{2}} \, r^{\frac{k}{2}} \, \mathrm{e}^{- \frac{r^2+\nu^2}{2}} \, I_{\frac{k}{2}-1}(\nu \, r) \\
&\propto r^{\frac{k}{2}} \, \mathrm{e}^{- \frac{r^2}{2}} \, I_{\frac{k}{2}-1}(\nu \, r) \label{eq:noncentralChi}
\end{align}
for $k=2$ (in this case, it is the Rician distribution as well).
Note that the order of the modified Bessel function is always $0$ in Eq.~\eqref{eq:pr} (i.e., $I_0$), while that of the noncentral chi distribution depends on the exponent of $r$.
This difference is more apparent by rewriting Eq.~\eqref{eq:noncentralChi} as
\begin{align}
\chi' (r;k,\nu) &= \sum_{n=0}^\infty \frac{{\rm e}^{-\frac{\nu^2}{2}}(\frac{\nu^2}{2})^n}{n!} \, \frac{r^{2n+k-1} \, {\rm e}^{-\frac{r^2}{2}}}{2^{n+\frac{k}{2}-1} \, \Gamma(n+\frac{k}{2})} \nonumber \\
&= \sum_{n=0}^\infty {\rm Poisson}\Bigl(n; \frac{\nu^2}{2}\Bigr) \, \chi\bigl(r;2n+k\bigr), \label{eq:chi_}
\end{align}
where ${\rm Poisson}(\,\cdot\,; \lambda)$ is the Poisson distribution, $\chi(\,\cdot\,;k)$ is the chi distribution, and $I_{\frac{k}{2}-1}(x)= \sum_{n=0}^\infty \frac{1}{n! \, \Gamma(n+\frac{k}{2})} (\frac{x}{2})^{2n+\frac{k}{2}-1}$ is used.
It indicates that the noncentral chi distribution is a mixture distribution of the chi distribution whose parameter $n$ is determined by the Poisson distribution.
On the other hand, Eq.~\eqref{eq:chi_like} can be rewritten as
\begin{align}
p(r;\nu,2,\frac{k}{2}) &= \sum_{n=0}^\infty \frac{(\frac{k}{2})_n \, (\frac{\nu^2}{2})^n}{n!^2 \, L_{-\frac{k}{2}}(\frac{\nu^2}{2})} \, \frac{r^{2n+k-1} \, {\rm e}^{-\frac{r^2}{2}}}{2^{n+\frac{k}{2}-1} \, \Gamma(n+\frac{k}{2})} \nonumber \\
&= \sum_{n=0}^\infty {\rm Poisson'}\Bigl(n; \frac{\nu^2}{2}, \frac{k}{2}\Bigr) \, \chi\bigl(r;2n+k\bigr),
\end{align}
where $(\cdot)_n$ is the Pochhammer symbol,
\begin{align}
{\rm Poisson'}(n; \lambda, \alpha) = \frac{(\alpha)_n \, \lambda^n}{n!^2 \, L_{-\alpha}(\lambda)}, \label{eq:poiss_}
\end{align}
and $I_0(x)= \sum_{n=0}^\infty \frac{1}{n! \, \Gamma(n+1)} (\frac{x}{2})^{2n}$ is used.
The modified Poisson distribution in Eq.~\eqref{eq:poiss_} is skewed according to the parameter $\alpha$, where $\alpha=1$ reduces to the Poisson distribution.
Therefore, the noncentral chi distribution and Eq.~\eqref{eq:pr} can be interpreted as the marginal distribution of a chi distribution whose degrees of freedom are incremented by a random non-negative integer obaying the discrete (Poisson or modified Poisson) distributions.

Furthermore, by writing the inverse variance (rate) and noncentrality as $\beta=\frac{1}{\sigma^2} \in \mathbb{R}_{>0}$ and $\lambda = \frac{\nu^2}{\sigma^2} = \beta \, |\mu|^2 \in \mathbb{R}_{\ge 0}$, respectively, and viewing as a distribution for power $x = r^2$, Eq.~\eqref{eq:pr} can be rewritten as
\begin{align}
p(x;\alpha,\beta,\lambda) &= \frac{\beta^{\alpha} }{\Gamma(\alpha) \, L_{-\alpha}(\lambda)} \, x^{\alpha-1} \, \mathrm{e}^{- \beta x} \, I_0(2\sqrt{\beta \, \lambda \, x}), \label{eq:gammap}
\end{align}
which reduces to the gamma distribution,
\begin{align}
{\rm Gamma}(x;\alpha,\beta) &= \frac{\beta^{\alpha} }{\Gamma(\alpha)} \, x^{\alpha-1} \, \mathrm{e}^{- \beta x}, \label{eq:gamma}
\end{align}
when $\lambda = 0$.
Therefore, Eq.~\eqref{eq:pr} is also a generalization of the gamma distribution.
In other words, the gamma distribution can be viewed as a distribution of magnitude of PW-NCCG centered at the origin ($\lambda = \nu = \mu = 0$).
Using the modified Poisson distribution in Eq.~\eqref{eq:poiss_}, Eq.~\eqref{eq:gammap} can be written as
\begin{align}
p(x;\alpha,\beta,\lambda)
= \sum_{n=0}^\infty {\rm Poisson'}(n; \lambda, \alpha) \, {\rm Gamma}(x;n+\alpha,\beta), \label{eq:gammap2}
\end{align}
which indicates that the proposed distribution can be interpreted as a mixture distribution of the gamma and modified Poisson distributions.
Although the distributions for magnitude in Eq.~\eqref{eq:pr} and for power in Eq.~\eqref{eq:gammap} are essentially the same, we will only consider Eq.~\eqref{eq:gammap} hereafter because of its simplicity.
Further statistical properties of this distribution, such as the mean, variance, and kurtosis, as well as detailed comparisons are provided in \cite{nakashika2026power}.

\section{PolarRBM and LogPolarRBM}
\label{sec:proposedRBMs}

Similar to standard BMs, the representational capacity of PolarBM can be significantly enhanced by introducing hidden units.
However, due to computational limitations regarding training and inference, we adopt a restricted connectivity structure analogous to that of RBMs (i.e., permitting connections only between visible and hidden units).

\subsection{PolarRBM}

According to PolarBM in Eq.~\eqref{eq:eulerbm1}, a natural choice for the distribution over complex-valued visible units $\bs{z} \in \mathbb{C}^D$ and binary hidden units $\bs{h} \in [0,1]^H$ can be given as
\begin{align}
p(\bs{z},\bs{h}) =& \frac{1}{Z}\mathrm{e}^{-E(\bs{z},\bs{h})}, \\
    E(\bs{z},\bs{h}) =& - \Re[\bs{z}^{\sf H} \mathbf{W} \bs{h]} 
    +\bs{a}^{\sf T}|\bs{z}|^2 - \Re \left[ \bs{b}^{\sf H} \bs{z} \right] - \bs{c}^{\sf T} \bs{h}, \label{eq:eulerrbm1}
\end{align}
where $Z = \int \sum_{\bs{h}} p(\bs{z},\bs{h}) \, \mathrm{d} \bs{z}$ is the normalization term, $\mathbf{W} \in \mathbb{C}^{D \times H}$ are the connection weights between $\bs{z}$ and $\bs{h}$, and $\bs{a} \in \mathbb{R}_{>0}^D$ and $\bs{c} \in \mathbb{R}^H$ are the bias parameters.
The conditional probability distributions of the visible and hidden units are given as follows:
\begin{align}
p(\bs{z}|\bs{h}) &= \mathcal{N}_{\mathrm{c}}\Bigl(\bs{z}; \frac{\mathbf{W}\bs{h}+\bs{b}}{2\bs{a}}, \frac{\bs{1}}{\bs{a}}\Bigr), \label{eq:defCondPDF(p(z|h))}\\
p(\bs{h}|\bs{z}) &= \mathcal{B}\bigl(\bs{h}; \bs{\varsigma}(\Re[\mathbf{W}^{\sf H}\bs{z}]+\bs{c})\bigr),
\end{align}
where the divisions in Eq.~\eqref{eq:defCondPDF(p(z|h))} are treated element-wise.
Although this model is naturally derived from PolarBM, the role of the hidden units is somewhat limited because they cannot control the squared magnitude of $\bs{z}$ (i.e., $|\bs{z}|^2$).

In this paper, we propose an improved version of Eq.~\eqref{eq:eulerrbm1} in which the hidden units control the variance, defined as follows:
\begin{align}
p(\bs{z},\bs{h}) =& \frac{1}{Z}\mathrm{e}^{-E(\bs{z},\bs{h})}, \\
    E(\bs{z},\bs{h}) =& |\bs{z}|^2{^{\sf T}}\mathbf{U}\bs{h}- \Re[\bs{z}^{\sf H} \mathbf{W} \bs{h]} 
    - \Re \left[ \bs{b}^{\sf H} \bs{z} \right] - \bs{c}^{\sf T} \bs{h} , \label{eq:eulerrbm2}
\end{align}
where $\mathbf{U} \in \mathbb{R}_{>0}^{D \times H}$ is the weight matrix.
The difference between Eqs.~\eqref{eq:eulerrbm1} and \eqref{eq:eulerrbm2} is the dependence of the inverse variance $\bs{a}$ $(=\mathbf{U}\bs{h})$ on $\bs{h}$.
The conditional probability distributions of the visible and hidden units for this model become
\begin{align}
p(\bs{z}|\bs{h}) &= \mathcal{N}_{\mathrm{c}}\Bigl(\bs{z}; \frac{\mathbf{W}\bs{h}+\bs{b}} {2\mathbf{U}\bs{h}}, \frac{\bs{1}}{\mathbf{U}\bs{h}}\Bigr), \label{eq:pz_h} \\
p(\bs{h}|\bs{z}) &= \mathcal{B}\bigl(\bs{h}; \bs{\varsigma}(-\mathbf{U}^{\sf T}|\bs{z}|^2+\Re[\mathbf{W}^{\sf H}\bs{z}]+\bs{c})\bigr).
\end{align}
We refer to the RBM defined by Eq.~\eqref{eq:eulerrbm2} as \textit{PolarRBM}.
As usual for standard RBMs, PolarRBM can be efficiently trained thanks to the conditional independence of each unit.

\subsection{LogPolarRBM}

Similar to PolarRBM in Eq.~\eqref{eq:eulerrbm2}, we propose an RBM version of LogPolarBM, termed \textit{LogPolarRBM}, by introducing hidden units $\bs{h} \in [0,1]^H$ as follows:
\begin{align}
p(\bs{z},\bs{h}) =& \frac{1}{Z}\mathrm{e}^{-E(\bs{z},\bs{h})}, \\
    E(\bs{z},\bs{h}) =& |\bs{z}|^2{^{\sf T}}\mathbf{U}\bs{h}- \Re[\bs{z}^{\sf H} \mathbf{W} \bs{h]} - \log|\bs{z}|{^{\sf T}} (\mathbf{V} \bs{h}-\bs{2}) \nonumber \\
    &- \Re \left[ \bs{b}^{\sf H} \bs{z} \right] - \bs{c}^{\sf T} \bs{h} , \label{eq:logpolarrbm}
\end{align}
where $Z = \int \sum_{\bs{h}} p(\bs{z},\bs{h}) \, \mathrm{d} \bs{z} $ is the normalization term. The parameters $\mathbf{W} \in \mathbb{C}^{D \times H}$, $\mathbf{V} \in \mathbb{R}_{>0}^{D \times H}$, and $\mathbf{U} \in \mathbb{R}_{>0}^{D \times H}$ represent the weight matrices connecting the hidden units and the complex visible units, their log-amplitudes, and their power, respectively. The vectors $\bs{b} \in \mathbb{C}^D$ and $\bs{c} \in \mathbb{R}^H$ denote the bias parameters for the visible and hidden units. Note that $-2$ in the third term of Eq.~\eqref{eq:logpolarrbm} is introduced to satisfy the shape parameter constraint $\alpha > 0$ of the PW-NCCG distribution. 

Under this definition, the conditional probability distributions for the visible and hidden units are derived as follows:
\begin{align}
p(\bs{z}|\bs{h}) &= \mathcal{N}_{\mathrm{pw}}\Bigl(\bs{z}; \frac{\mathbf{W}\bs{h}+\bs{b}} {2\mathbf{U}\bs{h}}, \frac{\bs{1}}{\mathbf{U}\bs{h}}, \frac{\mathbf{V}\bs{h}}{2}\Bigr), \label{eq:pz_h} \\
p(\bs{h}|\bs{z}) &= \mathcal{B}\bigl(\bs{h}; \bs{\varsigma}(-\mathbf{U}^{\sf T}|\bs{z}^2|+\Re[\mathbf{W}^{\sf H}\bs{z}]+\mathbf{V}^{\sf T}\log|\bs{z}|+\bs{c})\bigr).
\end{align}
When modeling speech spectra using LogPolarRBM, the binary hidden units $\bs{h}$ are expected to represent underlying information such as discriminative features, phonemes, or acoustic events. Eq.~\eqref{eq:pz_h} implies that, given such latent information, the spectra follow a PW-NCCG distribution.

In conventional audio and acoustic signal processing, as represented by the Local Gaussian Model (LGM), short-term spectra have frequently been modeled using a zero-mean, circularly-symmetric complex Gaussian distribution \cite{ephraim1984speech,kitamura2016determined,fevotte2009nonnegative,sawada2013multichannel}:
\begin{align}
p(\bs{z}) &= \mathcal{N}_{\mathrm{c}}(\bs{z}; 0,\sigma^2).
\end{align}
While the complex Gaussian assumption is statistically valid for long-term spectra, where the central limit theorem holds due to the mixture of various acoustic events, it is often inadequate for short-term spectra. In such cases, the signal is dominated by phoneme-specific deterministic features that do not cancel out like global noise. Instead, they introduce a local bias (non-zero mean) in the complex plane, which cannot be captured by the conventional zero-mean assumption.
In contrast, the proposed model in Eq.~\eqref{eq:pz_h} can appropriately characterize the shift of the mean from the origin in the complex domain, which arises from the deterministic structures of specific acoustic events (e.g., vowels), through its non-centrality parameters. Furthermore, since the model can control the heavy-tailed characteristics, it provides a more accurate and robust representation of short-term speech structures.

\subsection{Objective Function and Parameter Optimization}

While PolarBM and LogPolarBM can be trained similarly to standard BMs, we introduce a training procedure specifically for PolarRBM and LogPolarRBM for simplicity. Given a set of mutually independent data samples $\mathcal{Z}$ ($\ni\bs{z}$), the model parameters $\theta\in\{\mathbf{W}, \mathbf{U}, \bs{b}, \bs{c}\}$ for PolarRBM and $\theta\in\{\mathbf{W}, \mathbf{U}, \mathbf{V}, \bs{b}, \bs{c}\}$ for LogPolarRBM are both optimized by maximizing the following log-likelihood function:
\begin{align}
\mathcal{L}(\theta) = \mathbb{E}_{\bs{z}} \bigl[\,\log p(\bs{z})\,\bigr] = \mathbb{E}_{\bs{z}} \Bigl[\,\sum_{\bs{h}} p(\bs{z},\bs{h})\,\Bigr].
\label{eq:objectiveFunctionForTraining}
\end{align}
Since an analytical solution to this optimization problem is generally unavailable, we adopt the complex Adam (CAdam) optimizer \cite{nakashika2018complex}, which updates the model parameters as follows:
\begin{align}
    \theta^{(l+1)} \leftarrow \theta^{(l)} + 
    \alpha \, \Delta \theta^{(l)},
\end{align}
where $\alpha \in \mathbb{C}$ is a complex-valued learning rate satisfying $\Re[\alpha]>0$, the superscript $l$ represents the iteration counter, 
\begin{align}
\Delta \theta^{(l)} &= 2 \, \frac{1-\beta_2^l}{\sqrt{1-\beta_1^l}} \frac{m^{(l)}}{\sqrt{v^{(l)}}}, 
\label{eq:CAdam1}\\
m^{(l)} &= \beta_1 \, m^{(l-1)} + (1-\beta_1) \overline{\frac{\partial \mathcal{L}}{\partial \theta}}, \label{eq:CAdam2}\\
v^{(l)} &= \beta_2 \, v^{(l-1)} + (1-\beta_2) \left| \frac{\partial \mathcal{L}}{\partial \theta} \right|^2,
\label{eq:CAdam3}
\end{align}
and $\beta_1, \beta_2 \in \mathbb{R}$ $(0 < \beta_1, \beta_2 < 1)$ denote the moment decay rates of CAdam. The initial values for the first and second moments, $m^{(0)}$ and $v^{(0)}$, are set to zero.
The gradient $\partial \mathcal{L} / \partial \theta$ is computed using the Wirtinger derivative \cite{brandwood1983complex,kreutz2009complex}:
\begin{align}
\frac{\partial \mathcal{L}}{\partial \theta} = \frac{1}{2} \left( \frac{\partial \mathcal{L}}{\partial \Re[\theta]} - \mathrm{i} \frac{\partial \mathcal{L}}{\partial \Im[\theta]} \right).
\end{align}
The gradient of Eq.~\eqref{eq:objectiveFunctionForTraining} can be evaluated as follows:
\begin{align}
\frac{\partial \mathcal{L}}{\partial \theta} =  \mathbb{E}_{\bs{z}} \left[ -\frac{\partial E}{\partial \theta} \right]  - \mathbb{E}_{\bs{z},\bs{h}} \left[ -\frac{\partial E}{\partial \theta} \right], \label{eq:grad}
\end{align}
where the derivative for each parameter, $-\partial E/\partial \theta$, is given by
\begin{align}
- \frac{\partial E}{\partial \bs{b}} &= \frac{\overline{\bs{z}}}{2}, &
- \frac{\partial E}{\partial \bs{c}} &= \frac{\bs{h}}{2}, \\
- \frac{\partial E}{\partial \mathbf{U}} &= - \frac{|\bs{z}|^2 \, \bs{h}^{\sf T}}{2}, &
- \frac{\partial E}{\partial \mathbf{V}} &= \frac{\log|\bs{z}| \, \bs{h}^{\sf T}}{2}, \\
- \frac{\partial E}{\partial \mathbf{W}} &= \frac{\overline{\bs{z}} \, \bs{h}^{\sf T}}{2}.
\end{align}
For the parameters requiring strictly positive entries (i.e., $\mathbf{V}$ and $\mathbf{U}$), a potivity constraint is enforced using the softplus function~\cite{nakashika2021gamma}.

As shown in Eq.~\eqref{eq:grad}, the gradient of the log-likelihood function involves an expectation with respect to the model distribution, $\mathbb{E}_{\bs{z},\bs{h}} [\, -\partial E/\partial \theta \,]$, which is generally intractable to compute. Therefore, consistent with standard RBM training procedures, contrastive divergence (CD) is employed to approximate this derivative:
\begin{align}
\frac{\partial \mathcal{L}}{\partial \theta} \approx \mathbb{E}_{\bs{z}} \left[ -\frac{\partial E}{\partial \theta} \right]  - \mathbb{E}_{\widetilde{\bs{z}}} \left[ -\frac{\partial E}{\partial \theta} \right], 
\end{align}
where the expectation over the model distribution is approximated by the expectation over samples $\widetilde{\bs{z}}\sim p(\bs{z}|\hat{\bs{h}})$, with $\hat{\bs{h}} = \mathbb{E}_{p(\bs{h}|\bs{z})}[\,\bs{h}\,]$, generated from the training data $\bs{z}$.

\section{Experiments}
\label{sec:exp}

\subsection{Experimental Conditions}

To evaluate the effectiveness of the proposed method, speech representation experiments were conducted using speech data of a female announcer (\texttt{FTK}) in Set B of the ATR Japanese speech database.
A total of 50 utterances (approximately 4.2 minutes) were used for training. 
The original sampling rate of 20\,kHz was downsampled to 16\,kHz. 
Complex spectra obtained via a short-time Fourier transform (STFT) with a window length of 256 samples and an overlap of 64 samples served as the complex-valued visible units, yielding 129-dimensional complex spectral vectors.
The total number of frames was $49\,012$. 
PolarRBMs and LogPolarRBMs were trained and evaluated using varying numbers of hidden units.
The parameters of these models were optimized via stochastic gradient descent using the complex Adam optimizer \cite{nakashika2018complex} with a batch size of $100$, a learning rate of $0.01$, decay rates of $\beta_1 = 0.9$ and $\beta_2 = 0.999$, and $100$ training epochs.

For evaluation, complex spectra computed from an unseen test set of 53 utterances by the same speaker were encoded by calculating the conditional expectations of the hidden units given the visible units. 
The visible units were subsequently reconstructed by computing their conditional expectations from the hidden representations.
For LogPolarRBM, two reconstruction methods were investigated because the exact computation of the conditional expectation for the visible units is computationally expensive.
The first approach directly utilized the exact expectation (hereafter denoted as \texttt{LogPolarRBM}), whereas the second approximated the ratio of Laguerre functions (denoted as \texttt{LogPolarRBM(a)}) as detailed in Section~\ref{ssec:expect}.
The reconstructed visible units were then converted back into time-domain speech signals via an inverse STFT followed by overlap-add (OLA) synthesis.

The quality of the reconstructed speech was evaluated using both objective and subjective metrics.
The objective evaluation employed PESQ, STOI, UTMOS \cite{saeki2022utmos}, and MWARP-Q (Mapped WARP-Q) \cite{jassim2021warp}. 
For the subjective evaluation, the mean opinion score (MOS) was calculated from five-point ratings ranging from ``1: very poor'' to ``5: very good'' collected from 18 participants. 
Additionally, the reconstruction speed was evaluated using the real-time factor (RTF).

For comparison, conventional RBMs trained under the same conditions were used, including a complex-valued RBM (\texttt{CRBM}), a GVM-RBM \cite{nakashika2025gamma}, and a GaussRBM. 
Since GaussRBM accepts only real-valued inputs, two variants were implemented: \texttt{RBM-GL}, which utilized log-amplitude spectra as input features and reconstructed phase information using the Griffin--Lim algorithm, and \texttt{RBM-DUBM}, which simultaneously modeled phase information using a DUBM.
Furthermore, as reference methods, several deep-learning-based approaches (HiFi-GAN \cite{kong2020hifi}, CVAE \cite{nakashika2020complex}, and VAE) and a signal-processing-based approach (WORLD \cite{morise2016world}) were evaluated. 
The CVAE consisted of an encoder with dimensions [129-100-3$n$/2] and a decoder with dimensions [$n$/2-100-129] using the $\tanh$ activation function, where $n$ denotes the dimensionality of the latent features varied across the experiments. 
The VAE used concatenated real and imaginary components as input features, featuring an encoder with dimensions [258-200-2$n$] and a decoder with dimensions [$n$-200-258].
It should be noted that HiFi-GAN was included only as a reference system rather than a direct comparison target. 
The HiFi-GAN model reconstructed speech signals from 80-dimensional mel-spectrograms and was trained on approximately 26 hours of speech data from the JVS corpus.

\begin{table*}[t]
  \caption{Summary of objective and subjective evaluations. ``BM,'' ``DL,'' and ``SP'' denote Boltzmann machine, deep learning, and signal processing, respectively. ``CI'' indicates the 95\% confidence interval.}
  \vspace{5pt}
  \label{tab:res}
  \centering
  \begin{tabular}{llcccccc}
    \toprule
    \textbf{Type} & \textbf{Methods} & \textbf{PESQ}$\uparrow$ & \textbf{STOI}$\uparrow$ & \textbf{UTMOS}$\uparrow$ &\textbf{MWARP-Q}$\uparrow$ &\textbf{RTF}$\downarrow$ &\textbf{MOS$\uparrow$ {\scriptsize($\pm$CI)}}\\
    \midrule
    \multirow{7}{*}{\textbf{BM-based}} & \textbf{LogPolarRBM} & $\bs{4.00}$& $\bs{0.995}$&$\underline{3.92}$ &$\bs{4.69}$  &$86.3$ &$\!\underline{4.51}${\scriptsize$\pm 0.16\!\!$}\\
    &\textbf{LogPolarRBM(a)} & $\underline{3.93}$& $\bs{0.995}$&$\bs{3.93}$ &$\bs{4.69}$  &$0.12$ &$\!\bs{4.65}${\scriptsize$\pm 0.13\!\!$}\\
    & \textbf{PolarRBM} & $3.65$& $\underline{0.994}$&$3.62$ &$3.10$  &$0.07$ &$3.38${\scriptsize$\pm 0.24$}\\
    & \textbf{GVM-RBM \cite{nakashika2025gamma}} & $3.75$& $0.986$&$3.79$ &$3.30$  &$0.10$ &$4.28${\scriptsize$\pm 0.15$}\\
    & \textbf{CRBM \cite{nakashika2018complex}} & $2.70$& $0.944$&$2.65$ &$2.88$ &$\underline{0.06}$ &$3.14${\scriptsize$\pm0.22$} \\
    & \textbf{RBM-DUBM}\hspace{-8pt} & $3.17$& $0.952$&$2.76$ &$\underline{3.48}$  &$0.08$ &$2.96${\scriptsize$\pm0.17$} \\
    & \textbf{RBM-GL}\hspace{-8pt} & $2.62$& $0.950$&$2.26$ &$3.19$  &$0.48$ &$2.17${\scriptsize$\pm0.19$} \\    \midrule
    \multirow{3}{*}{\textbf{DL-based}} & \textbf{HiFi-GAN \cite{kong2020hifi}} & $3.32$ & $0.956$&$3.52$  &$3.00$  &$0.27$ &$4.21${\scriptsize$\pm0.17$} \\
    & \textbf{CVAE \cite{nakashika2020complex}} & $3.48$& $0.987$&$3.67$ &$3.28$ &$\bs{0.01}$ &$3.95${\scriptsize$\pm0.18$} \\
    & \textbf{VAE} & $2.46$& $0.938$&$1.97$ &$2.96$  &$\bs{0.01}$ &$1.57${\scriptsize$\pm0.17$} \\
    \midrule
    \textbf{SP-based} & \textbf{WORLD \cite{morise2016world}}& $2.97$& $0.961$&$2.50$ &$2.51$  &$0.44$ &$3.34${\scriptsize$\pm0.23$} \\
    \midrule
    & \textbf{NATURAL}& ---  & ---  &$3.94$ & ---  & --- &$4.43${\scriptsize$\pm0.14$} \\
    \bottomrule
  \end{tabular}
\end{table*}

\subsection{Expectation of Visible Units of LogPolarRBM}
\label{ssec:expect}

When reconstructing the visible units $\bs{z}$ from the hidden units $\bs{h}$, it is standard practice to use the conditional expectation $\mathbb{E}_{p(\bs{z}|\bs{h})}[\bs{z}]$ as the reconstruction.
For LogPolarRBM, the conditional distribution $p(\bs{z}|\bs{h})$ follows the PW-NCCG distribution as shown in Eq.~\eqref{eq:pz_h}. 
Since the visible units are conditionally independent, we consider the expectation of a single visible unit $z$ without loss of generality. 
Given the corresponding PW-NCCG parameters $\mu$, $\sigma^2$, and $\alpha$, the conditional expectation can be calculated as
\begin{align}
\mathbb{E}[z] = \mu \, R_{\alpha\!}\biggl(\frac{|\mu|^2}{\sigma^2}\biggr),
\qquad
R_\alpha(\lambda) =\frac{L_{\alpha-1}^{(1)}(-\lambda)}
     {L_{\alpha-1}(-\lambda)},
\end{align}
where $L_n^{(m)}(\cdot)$ denotes the generalized Laguerre function of order $n$ and degree $m$ (see Appendix~\ref{appendix:expect} for the detailed derivation).
However, evaluating this expectation is computationally expensive because it involves the Laguerre-function ratio $R_\alpha(\lambda)$.
To address this computational bottleneck, we introduce an efficient approximation,
$\tilde{R}_\alpha(\lambda)$.

First, we analyze the boundary behavior of $R_\alpha(\lambda)$ for a fixed $\alpha$.
Using $L_n^{(m)}(0) = \frac{\Gamma(n+m+1)}{\Gamma(n+1)\,\Gamma(m+1)}$, we obtain
\begin{align}
R_\alpha(0) = \frac{L_{\alpha-1}^{(1)}(0)}{L_{\alpha-1}(0)}  = \alpha.
\label{eq:R0}
\end{align}
As $\lambda \to \infty$, the generalized Laguerre function exhibits the following asymptotic behavior: 
\begin{align}
L_n^{(m)}(-\lambda) \,\sim\, \frac{\lambda^n}{\Gamma(n+1)} \left(1+\frac{n(n+m)}{\lambda} \right), 
\end{align}
which yields the asymptotic approximation
\begin{align}
R_\alpha(\lambda) \approx 1+\frac{\alpha-1}{\lambda},
\label{eq:Rapprox1}
\end{align}
where the derivation is given in the Appendix~\ref{appendix:approx}.
Consequently,
\begin{align}
\lim_{\lambda\to\infty} R_\alpha(\lambda) = 1.
\label{eq:Rinf}
\end{align}

\begin{figure}[t]
    \centering
    \subfloat[Target function $R_\alpha(\lambda)$]{\label{fig:r_exact}\includegraphics[width=0.5\columnwidth]{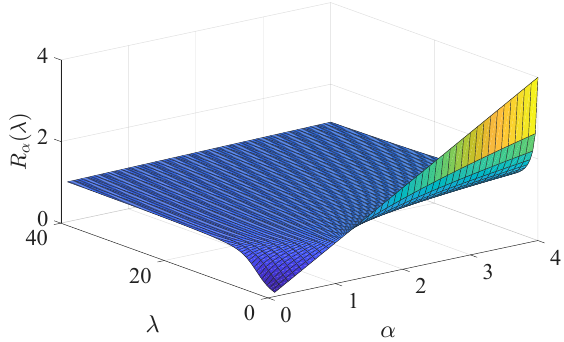}}
    \hfill
    \subfloat[Proposed approximation $\tilde{R}_\alpha(\lambda)$]{\label{fig:r_approx}\includegraphics[width=0.5\columnwidth]{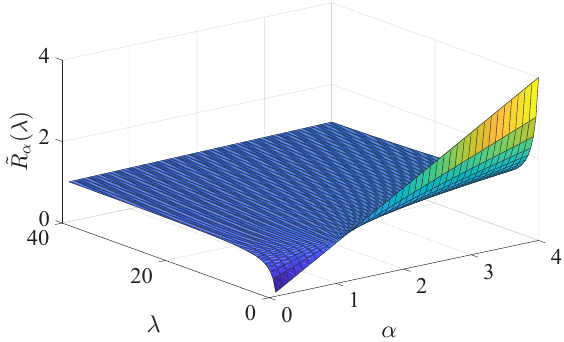}}
    \caption{Comparison between the exact Laguerre-function ratio $R_\alpha(\lambda)$ and its approximation $\tilde{R}_\alpha(\lambda)$.}
    \label{fig:r_comp}
\end{figure}

As a function satisfying Eqs.~\eqref{eq:R0} and \eqref{eq:Rinf} (i.e., $R_\alpha(0) = \alpha$ and $\lim_{\lambda\to\infty} R_\alpha(\lambda) = 1$), we propose the following approximation of the Laguerre-function ratio $R_\alpha(\lambda)$:
\begin{align}
\tilde{R}_\alpha(\lambda) = 1+\frac{\alpha-1}{1+\lambda} = \frac{\alpha+\lambda}{1+\lambda},
\label{eq:Rapprox2}
\end{align}
which is a slight modification of Eq.~\eqref{eq:Rapprox1}.
Fig.~\ref{fig:r_comp} visually compares these functions.
As illustrated, the proposed approximation closely tracks the exact ratio over a wide range of parameter values.
Specifically, the exact ratio monotonically increases toward $1$ for $\alpha<1$, is identically equal to $1$ for $\alpha=1$, and monotonically decreases toward $1$ for $\alpha>1$.
The proposed approximation effectively captures these characteristics at a significantly lower computational cost.

\begin{figure}[t]
\centerline{\includegraphics[width=1.0\columnwidth]{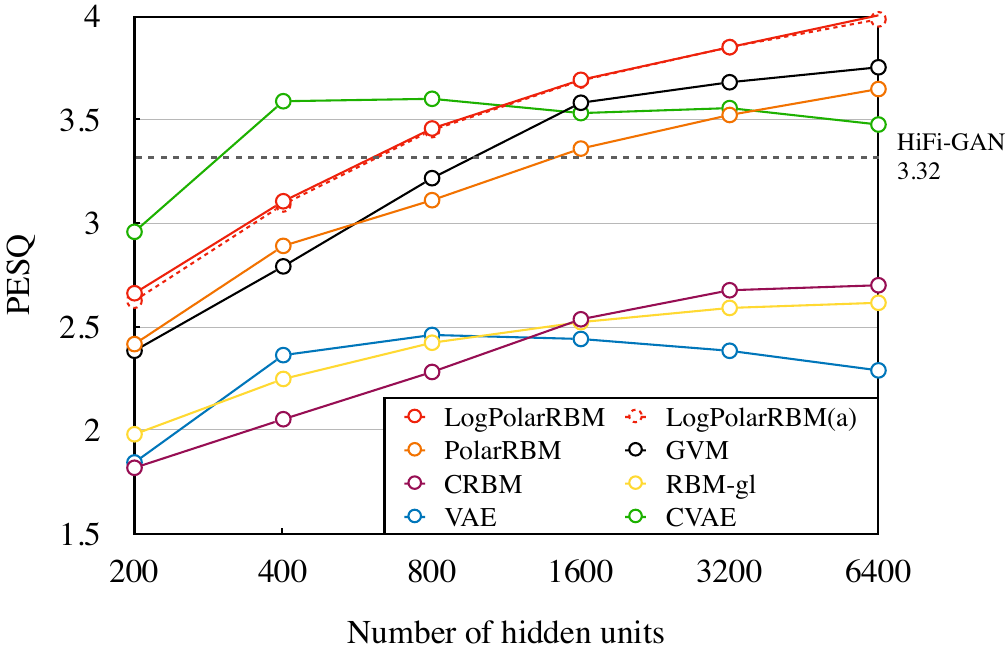}}
\caption{PESQ scores of each method across various numbers of hidden units.}
\label{fig:pesq_comp}
\end{figure}

\subsection{Results and Discussion}

First, the performance of each method was evaluated across varying numbers of hidden units or latent variables, as illustrated in Fig.~\ref{fig:pesq_comp}. 
As shown in the figure, the PESQ scores of the BM-based methods increased with the number of hidden units.
In contrast, the VAE-based methods achieve their peak performance at a latent dimensionality of approximately 800, beyond which their performance gradually degrades.
Among the compared methods, LogPolarRBM achieves the highest overall performance.
Although the approximated variant, \texttt{LogPolarRBM(a)}, exhibits a slight performance degradation compared to the exact \texttt{LogPolarRBM}, this difference remains marginal. 
LogPolarRBM, PolarRBM, and GVM significantly outperform the other BMs that rely on Cartesian representations, demonstrating the efficacy of modeling complex spectra in polar coordinates.
Notably, PolarRBM achieves performance comparable to GVM despite utilizing a simple complex Gaussian distribution without logarithmic modeling.

Table~\ref{tab:res} summarizes the speech reconstruction quality metrics for each method.
The results confirm that PolarRBM outperforms conventional RBMs across multiple evaluation metrics, particularly CRBM and RBM-DUBM, both of which are explicitly designed to represent complex-valued structures.
This outcome indicates that polar coordinates enable PolarRBM to capture complex structures more effectively.
Furthermore, LogPolarRBM, which incorporates logarithmic modeling, achieves the highest performance across nearly all metrics.
LogPolarRBM and LogPolarRBM(a) exhibit nearly identical performance across all evaluation metrics except for the RTF.
While the exact LogPolarRBM yields an RTF of 86.3, the approximated variant LogPolarRBM(a) achieves an RTF of 0.12.
These results indicate that the proposed approximation maintains reconstruction quality while significantly improving computational efficiency, rendering LogPolarRBM(a) a more practical choice.
Both the objective and subjective evaluation results demonstrate that the proposed methods outperform state-of-the-art deep-learning-based approaches.
This finding is particularly noteworthy given that the proposed RBMs are shallow architectures with significantly lower structural complexity than deep networks.

Remarkably, according to Welch's $t$-test at the 5\% significance level, the subjective evaluations of the proposed models (LogPolarRBM and LogPolarRBM(a)) and the natural speech (NATURAL) had no statistically significant difference.
Moreover, the average MOS values for the proposed methods are slightly higher than those of NATURAL.
These results suggest that the proposed methods achieve a perceptual naturalness comparable to, or marginally exceeding, that of the original speech.
A potential explanation is that the encoding-decoding process suppresses low-level noise, yielding speech that is perceived as cleaner while preserving its inherent naturalness.

\section{Conclusion}
\label{sec:conclusion}

In this study, we proposed a novel Boltzmann machine, termed PolarBM, which models interactions among complex-valued units in polar coordinates, along with its restricted variant, PolarRBM. 
Furthermore, we proposed LogPolarBM and LogPolarRBM, which incorporate interactions in the log-amplitude domain to better align with human auditory perception. 
We demonstrated that these models yield the PW-NCCG distribution capable of characterizing a diverse range of amplitude distributions.
Speech representation experiments demonstrated that the proposed PolarRBM achieved a reconstruction quality comparable or superior to that of conventional RBMs and state-of-the-art deep learning methods across both objective and subjective evaluations.
Moreover, LogPolarRBM further enhanced reconstruction quality, consistently outperforming existing approaches.
In particular, LogPolarRBM achieved a perceptual quality comparable to that of natural speech, with no statistically significant differences observed in the subjective evaluations.
Although exactly computing the visible-unit expectation in LogPolarRBM is computationally demanding, the proposed approximation achieves nearly identical performance while significantly reducing computational cost, making the model practical for real-world applications.
Future work includes integrating these proposed models into deep neural architectures and applying them to a broader range of speech-processing tasks, such as automatic speech recognition, speech synthesis, and source separation. 
Additionally, applications to other domains that involve complex-valued data, including image processing and communication systems, presents a promising direction for future research.

\appendices

\section{Derivation of the Expectation of PW-NCCG}
\label{appendix:expect}

The expectation of a PW-NCCG random variable is derived as follows.
The expectation of $z \sim \mathcal{N}_{\mathrm{pw}}\bigl(z; \mu, \sigma^2, \alpha\bigr)$ in polar coordinate, $z = r \, {\rm e}^{{\rm i}\theta}$, is given by
\begin{align}
\mathbb{E}[z] &= \int_\mathbb{C} z \, p(z) \, \mathrm{d}z \\
&= \int_0^\infty \int_{-\pi}^\pi r \, {\rm e}^{\rm i \theta} \, p(r,\theta) \, r \, \mathrm{d}r \, \mathrm{d}\theta \\
&= \frac{{\rm e}^{-\frac{|\mu|^2}{\sigma^2}}}{\pi \, \sigma^{2\alpha} \, \Gamma(\alpha) \, L_{\alpha-1}(-\frac{|\mu|^2}{\sigma^2})} \int_0^\infty r^{2\alpha} \, {\rm e}^{-\frac{r^2}{\sigma^2}} \notag \\
&\hspace{20mm}\times \left[ \int_{-\pi}^\pi {\rm e}^{{\rm i} \theta + \frac{2 |\mu| r }{\sigma^2} \cos(\theta - \angle \mu)} \, \mathrm{d}\theta \right] \mathrm{d}r.
\end{align}
The inner integral can be evaluated using the integral representation of the modified Bessel function of the first kind of order 1, $I_1(x)$:
\begin{align}
\int_{-\pi}^{\pi} {\rm e}^{{\rm i}\theta + x \cos(\theta-\phi)} \, \mathrm{d}\theta = 2\pi \, {\rm e}^{{\rm i}\phi} \, I_1(x),
\end{align}
which yields
\begin{align}
\mathbb{E}[z] = \frac{2 \, {\rm e}^{{\rm i}\angle\mu} \, {\rm e}^{-\frac{|\mu|^2}{\sigma^2}}}{\sigma^{2\alpha} \, \Gamma(\alpha) \, L_{\alpha-1}\!\left(-\frac{|\mu|^2}{\sigma^2}\right)} \int_0^\infty \!r^{2\alpha} \, {\rm e}^{-\frac{r^2}{\sigma^2}} \, I_{1\!}\biggl(\frac{2\,r\,|\mu|}{\sigma^2}\biggr) \, \mathrm{d}r.
\end{align}
Substituting the power-series expansion of the modified Bessel function
\begin{align}
I_1(x) = \sum_{k=0}^{\infty} \frac{1}{k!\,\Gamma(k+2)} \left( \frac{x}{2} \right)^{2k+1},
\end{align}
and exchanging the order of summation and integration gives
\begin{align}
\mathbb{E}[z] &= \frac{2 \, {\rm e}^{{\rm i} \angle \mu} \, {\rm e}^{-\frac{|\mu|^2}{\sigma^2}}}{\sigma^{2\alpha} \, \Gamma(\alpha) \, L_{\alpha-1}(-\frac{|\mu|^2}{\sigma^2})} \sum_{k=0}^\infty \frac{|\mu|^{2k+1}}{k! \, \Gamma(k+2) \, \sigma^{4k+2}} \notag \\
&\hspace{20mm}\times \int_0^\infty r^{2(\alpha+k)+1} \, {\rm e}^{-\frac{r^2}{\sigma^2}} \, \mathrm{d}r .
\end{align}
The remaining integral is evaluated using
\begin{align}
\int_0^\infty r^{2a+1} \, {\rm e}^{-\frac{r^2}{\sigma^2}} \, \mathrm{d}r = \frac{\sigma^{2a+2}}{2} \Gamma(a+1),
\end{align}
which leads to
\begin{align}
\mathbb{E}[z] = \frac{\alpha \, \mu \, {\rm e}^{-\frac{|\mu|^2}{\sigma^2}}}{L_{\alpha-1}(-\frac{|\mu|^2}{\sigma^2})} \, {}_1F_1 \!\left(\alpha+1; 2; \frac{|\mu|^2}{\sigma^2} \right) .
\end{align}
By applying Kummer's transformation and the relation between the confluent hypergeometric function and the generalized Laguerre function, we finally obtain 
\begin{align}
\mathbb{E}[z] &= \mu \, \frac{ L_{\alpha-1}^{(1)} \!\left( -\frac{|\mu|^2}{\sigma^2} \right) }{ L_{\alpha-1} \!\left( -\frac{|\mu|^2}{\sigma^2} \right) } \\
&= \mu \, R_\alpha\!\left(\frac{|\mu|^2}{\sigma^2}\right), \qquad R_\alpha(\lambda) =\frac{L_{\alpha-1}^{(1)}(-\lambda)}{L_{\alpha-1}(-\lambda)}.
\end{align}

\section{Asymptotic Approximation of the Laguerre-Function Ratio}
\label{appendix:approx}

The asymptotic approximation of the Laguerre-function ratio
\begin{align}
R_\alpha(\lambda) =\frac{L_{\alpha-1}^{(1)}(-\lambda)}{L_{\alpha-1}(-\lambda)}
\end{align}
is derived as follows.
Using the asymptotic expansion of the generalized Laguerre function
\begin{align}
L_{n}^{(m)}(-\lambda) \,\sim\, \frac{\lambda^n}{\Gamma(n+1)} \left( 1 + \frac{n(n+m)}{\lambda}\right),
\end{align}
the following expression is obtained:
\begin{align}
R_\alpha(\lambda) \,\sim\, \frac{ 1+\frac{\alpha(\alpha-1)}{\lambda} }{ 1+\frac{(\alpha-1)^2}{\lambda}}. \label{eq:asympRelation}
\end{align}
Applying the expansion
\begin{align}
\frac{1}{1+x} = 1-x+O(x^2), \qquad x\to0,
\end{align}
to Eq.~\eqref{eq:asympRelation} gives
\begin{align}
R_\alpha(\lambda) \,&\sim\, \left( 1+\frac{\alpha(\alpha-1)}{\lambda} \right) \left( 1-\frac{(\alpha-1)^2}{\lambda} +O(\lambda^{-2}) \right) \\
&= 1 + \frac{\alpha(\alpha-1)-(\alpha-1)^2}{\lambda} + O(\lambda^{-2}) \\
&= 1+\frac{\alpha-1}{\lambda} +O(\lambda^{-2}).
\end{align}
Therefore, the Laguerre-function ratio can be expressed as
\begin{align}
R_\alpha(\lambda) = 1+\frac{\alpha-1}{\lambda} + O(\lambda^{-2}),
\qquad \lambda \to \infty.
\end{align}
Neglecting the higher-order terms yields the approximation
\begin{align}
R_\alpha(\lambda) \approx 1+\frac{\alpha-1}{\lambda},
\end{align}
which forms the basis of the proposed approximation.

\bibliography{refs}
\bibliographystyle{IEEEtran} 




\begin{IEEEbiography}{Toru Nakashika}
received his B.E. and M.E.degrees in computer science from Kobe University in 2009 and 2011, respectively.
On the summer in 2010, he was a student researcher at IBM Research, Tokyo Research Laboratory.
From September 2011 to August 2012, he was a visiting researcher in the image group at INSA de Lyon in France.
In the same year, he continued his research as a doctoral student at Kobe University, and received his Dr.Eng. degree in computer science in 2014.
To April 2015, he was an assistant professor at Kobe University.
In 2015, he joined the University of Electro-Communications as an assistant professor.
He is currently an associate professor of the Graduate School of Informatics and Engineering, the University of Electro-Communications.
He received the Young Researcher’s Award in IEICE Speech Field in 2013, the Best Paper Award in SIGMUS Ongaku Symposium 2016, the 44th Awaya Prize Young Researcher Award from the Acoustical Society of Japan, the 15th Itakura Prize Innovative Young Researcher Award from the Acoustical Society of Japan.
He is a member of the IEEE, the IEICE, the ASJ, the JSAI, and the ISCA.
\end{IEEEbiography}

\begin{IEEEbiography}{Kohei Yatabe}
received his B.E., M.E., and Ph.D. degrees from Waseda University in 2012, 2014, and 2017, respectively. He was an Assistant Professor of Waseda University from 2017 to 2022. He is currently an Associate Professor of Tokyo University of Agriculture and Technology.
\end{IEEEbiography}

\vfill




\end{document}